\documentclass[10pt,twocolumn,letterpaper]{article}
\usepackage[pagenumbers]{cvpr}

\usepackage{cuted} %

\usepackage{currfile} %

\usepackage{caption} %

\usepackage{amsmath,amssymb,bm}
\usepackage{booktabs}
\usepackage{multirow}
\usepackage{makecell}
\usepackage{xspace}
\usepackage{enumitem}
\usepackage{pifont}

\definecolor{cvprblue}{rgb}{0.21,0.49,0.74}
\usepackage[pagebackref,breaklinks,colorlinks,allcolors=cvprblue]{hyperref}
\hypersetup{pdftitle={Feed-Forward Adaptive Gaussian Splatting for Interactive Medical Volume Rendering},pdfauthor={Zhongpai Gao, Benjamin Planche, Meng Zheng, Anwesa Choudhuri, Chaoyi Zhou, Terrence Chen, Ziyan Wu}}

\def\confName{CVPR}
\def\confYear{2027}

\newcommand{\method}{RenderFM-Next\xspace}

\newcommand{\pmci}[2]{\scriptsize{[#1,\,#2]}}
\newcommand{\mainref}[1]{\cref{#1} of the main paper}

\title{Feed-Forward Adaptive Gaussian Splatting\\for Interactive Medical Volume Rendering}

\author{
Zhongpai Gao$^{1,*}$\hspace{0.8em}Benjamin Planche$^{1}$\hspace{0.8em}Meng Zheng$^{1}$\hspace{0.8em}Anwesa Choudhuri$^{1}$\\
Chaoyi Zhou$^{1,2,\dag}$\hspace{0.8em}Terrence Chen$^{1}$\hspace{0.8em}Ziyan Wu$^{1}$\\[4pt]
$^{1}$United Imaging Intelligence, Boston MA, USA\qquad $^{2}$Clemson University, Clemson SC, USA
}

\begin{document}
\maketitle
\begingroup
\renewcommand\thefootnote{}
\footnotetext{\textsuperscript{*}Corresponding author.\quad
\textsuperscript{\dag}The contribution of this author was made while interning at United Imaging Intelligence.}
\addtocounter{footnote}{-1}
\endgroup
\begin{strip}\centering
\vspace*{-18pt}
\includegraphics[width=\textwidth]{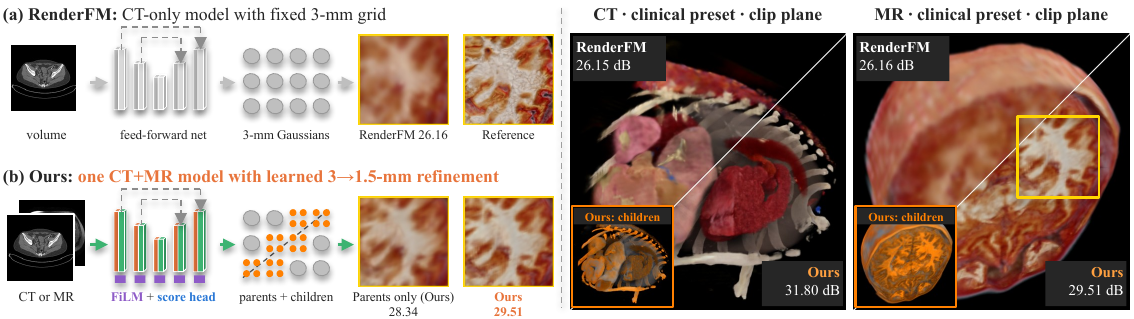}
\captionof{figure}{\textbf{Adaptive refinement for feed-forward medical Gaussians.} (a)~RenderFM~\cite{renderfm} uses a fixed
3-mm grid, which blurs organ boundaries and cut faces. (b)~\method, one CT+MR model, refines to 1.5-mm children
(orange) where a learned score expects the most benefit, under a user-set budget. Right: clipped clinical presets
(validation; RenderFM retrained; PSNR in dB).}
\label{fig:teaser}
\end{strip}

\begin{abstract}
Path-traced rendering shows CT and MR volumes with realistic lighting, but it restarts after every change of view,
clipping plane or transfer function (TF), the map from voxel intensity to colour and opacity, which limits the
frame rate.
Feed-forward Gaussian splatting models such as RenderFM predict the Gaussians directly from the volume instead of
fitting them to each scan, and are fast. RenderFM, however, handles only CT, runs its network again after every TF
edit, has no clipping operator, and places its Gaussians on a uniform 3-mm grid that blurs organ boundaries and cut
faces.
We present \method, a feed-forward model that addresses these limits.
One network handles both CT and MR. Its Gaussians are predicted once per volume: a new TF needs no pass of this network, and
a clipping plane cuts them exactly, using operators from prior work.
The model predicts coarse 3-mm Gaussians for the whole volume and refines them to 1.5\,mm only where refinement
improves the image. A small score head ranks the coarse Gaussians by how much refining them would reduce the
rendering error; at render time the user chooses how many are refined, and those next to a clipping plane are always
refined, which sharpens cut faces.
We build a benchmark of about 2,000 CT and MR scans with path-traced references under clinical presets, TF edits
and clipping.
Our \method improves over RenderFM retrained on this benchmark by 2.8\,dB on CT and 2.4\,dB on MR and renders a
$1600^2$ view in 4--6\,ms.
Refinement adds 0.30 / 0.69\,dB over the same model trained without it (0.51 / 1.17\,dB on cut faces), and refining
a learned 10\% of the coarse Gaussians matches a random 50\% with 2.4$\times$ fewer Gaussians.
Project page: \url{https://gaozhongpai.github.io/renderfmnext/}.
\end{abstract}

\section{Introduction}
\label{sec:intro}

Path-traced (``cinematic'') rendering of CT and MR volumes conveys depth, occlusion and tissue appearance through
global illumination~\cite{comaniciu2016shaping,dappa2016cinematic,eid2017cinematic}. Its image is a Monte Carlo
estimate that restarts whenever the viewpoint, a clipping plane or the transfer function (TF), which maps voxel
intensities to colour and opacity, changes~\cite{kroes2012exposure,novak2018monte}; even a single denoised pass takes tens of milliseconds, which limits
the frame rate for high-refresh and stereo displays.
These changes are frequent in practice: a reader switches between TFs that show bone, vessels or soft tissue, and
moves a clipping plane to look inside the body.
Gaussian splatting~\cite{kerbl3dgs} can stand in for the path tracer during interaction. The Gaussians can be fitted
to each scan, which takes minutes~\cite{niedermayr2024application,dyken2026veg,tang2025ivrgs}, or predicted from the
volume by a network in a single pass~\cite{renderfm}, within seconds for a new patient.

RenderFM~\cite{renderfm} takes the second route: a 3D U-Net maps a CT volume to one Gaussian per cell of a 3-mm grid.
This design has four limits (\cref{tab:latency}). The model is trained for CT only. The TF is a network input, so
the network runs again after every TF edit. There is no clipping operator, so a clipping plane can only keep or
remove whole Gaussians. And a uniform grid spends the same number of Gaussians on smooth tissue as on thin
structures: at 3\,mm, organ boundaries and the cut faces exposed by a clipping plane look blurred
(\cref{fig:teaser}), while a 1.5-mm grid would need eight times as many Gaussians.

We present \method, which addresses these limits.
One network takes CT or MR as input and is told which one through a learned modality embedding.
Its Gaussians depend on neither the TF nor the clipping plane: a new TF needs no pass of this network, and a clipping plane
cuts the Gaussians exactly.
To sharpen the image without a fine grid everywhere, the network predicts Gaussians at two resolutions: a coarse
Gaussian, the \emph{parent}, for every 3-mm cell, and up to eight 1.5-mm \emph{children} that can replace it.
At render time, a budget sets the fraction of parents that are replaced by their children, so the same model can
trade quality against speed without retraining. Parents next to a clipping plane are always refined, which sharpens
cut faces.

The key question is where to refine. Learning this decision together with the renderer lowered image quality in our
experiments (\cref{sec:exp:ablation}), so we separate the two.
We first train the renderer with random subsets of parents refined, so that it works at any budget.
We then freeze it, estimate for each parent how much the training loss would drop if it were refined, and train a
small score head to predict this benefit from the network's features.
At render time, the parents with the highest scores are refined.
For TF-editable appearance and for clipping we reuse two existing components,
FactorSplat~\cite{factorsplat} and XClipGS~\cite{xclipgs}.

We evaluate on a new benchmark of CT and MR scans with path-traced references. \method improves over a retrained
RenderFM by 2.8\,dB on CT and 2.4\,dB on MR and renders a $1600^2$ view in 4--6\,ms. Most of this gain is already
reached with parents only; refinement adds 0.30 / 0.69\,dB on CT / MR, and up to 1.17\,dB on cut faces.

Our contributions are:
\begin{itemize}[leftmargin=*,itemsep=1pt,topsep=2pt]
\item \textbf{One feed-forward model for CT and MR.} Its Gaussians are predicted once per volume and depend on
  neither the TF nor the clipping plane. With as many Gaussians as RenderFM, it is already better by 2.5\,dB on CT and by 1.7\,dB on MR (\cref{sec:exp:main}).
\item \textbf{Refinement with a render-time budget.} The network also predicts 1.5-mm children and is trained so that
  any fraction of the parents can be refined. A score head trained on the frozen renderer predicts which parents
  benefit most (\cref{sec:method:score}): refining a learned 10\% matches refining a random 50\% with
  2.4$\times$ fewer Gaussians (\cref{fig:quality_count}).
\item \textbf{A CT+MR rendering benchmark.} 1,008 CT and 958 MR scans from public sources with $1600^2$ path-traced
  references under clinical presets, TF edits and clipping, evaluated per patient with confidence
  intervals (\cref{sec:benchmark}).
\end{itemize}

\section{Related Work}
\label{sec:related}
\begin{table}[t]
\centering
\caption{\textbf{RenderFM and \method side by side.} Top: capabilities. Bottom: quality (\cref{tab:main}) and cost
(\cref{sec:exp:cost}) on CT / MR. $\rho$: fraction of parents refined.}
\label{tab:latency}
\setlength{\tabcolsep}{2.4pt}
\footnotesize
\begin{tabular}{@{}lccc@{}}
\toprule
 & & \multicolumn{2}{c}{\method\ (Ours)} \\
\cmidrule(l){3-4}
 & RenderFM~\cite{renderfm} & Parents only & \makecell{Children\\($\rho{=}20\%$)} \\
\midrule
CT and MR in one network & -- & \checkmark & \checkmark \\
New TF, no U-Net pass & -- & \checkmark & \checkmark \\
Exact clipping-plane cut & -- & \checkmark & \checkmark \\
Adjustable refinement & -- & -- & \checkmark \\
Refined cut faces & -- & -- & \checkmark \\
\midrule
PSNR (dB)$\uparrow$ & 26.87 / 26.25 & 29.39 / 27.95 & 29.69 / 28.63 \\
Frame rate (FPS)$\uparrow$ & 99 / 159 & 225 / 350 & 162 / 229 \\
New TF (ms)$\downarrow$ & 71 / 34 & 21 / 15 & 51 / 35 \\
New volume (s)$\downarrow$ & 1.99 / 1.56 & 0.82 / 0.54 & 1.73 / 1.06 \\
Peak memory (GB)$\downarrow$ & 3.3 / 1.7 & 3.0 / 1.5 & 7.7 / 4.9 \\
\bottomrule
\end{tabular}
\end{table}

\noindent\textbf{Feed-forward Gaussian splatting.}
Feed-forward models predict Gaussians in a single network pass. Some predict one Gaussian per input
pixel~\cite{splatterimage,pixelsplat,mvsplat}, and large reconstruction
models~\cite{lrm,gslrm,tang2024lgm,xu2024grm} predict a scene from a few images.
VolSplat~\cite{volsplat} predicts voxel-aligned Gaussians, and SparseSplat~\cite{sparsesplat} samples them by local
entropy. VVGT~\cite{vvgt} maps scientific volumes and rendered views to Gaussians without TF conditioning, and
MedGSSR~\cite{medgssr} predicts Gaussians from medical volumes for super-resolution.
Closest to our work, RenderFM~\cite{renderfm} uses an nnU-Net-style~\cite{nnunet} network to predict one
6D Gaussian~\cite{gao20246dgs} per cell of a fixed 3-mm grid from a CT volume, with the TF and an anatomy mask as
additional inputs.
Our \method keeps the volume-to-Gaussians design but makes the geometry independent of the TF, adds a second, finer
resolution, and handles CT and MR with one network (\cref{tab:latency} lists
all differences).

\noindent\textbf{Deciding where to place Gaussians.}
Per-scene 3DGS adapts the density of Gaussians during optimisation: by splitting where gradients are
large~\cite{kerbl3dgs,absgs,bulo2024revising}, relocation~\cite{mcmc-gs}, densification under a
budget~\cite{taming3dgs}, pruning~\cite{lightgaussian} and hierarchies~\cite{scaffold-gs,octree-gs}.
Among feed-forward models, ATSplat~\cite{atsplat} expands uncertain anchor tokens and
OctreeSplat~\cite{octreesplat} predicts octree densification scores. For cinematic CT,
Xiang \etal~\cite{wheregaussians} learn Gaussian positions under a budget from the volume, labels and TF, and then
fit the remaining attributes per patient; Martin \etal~\cite{martin2026fixedbudget} place a fixed budget
analytically from local field structure and refine it per volume.
Our \method predicts all attributes at both resolutions without per-patient fitting, supports a range of budgets with
one model, and learns its refinement score from the measured effect of refinement on the rendering loss.

\noindent\textbf{Medical volume rendering with Gaussians.}
Direct volume rendering integrates emission and absorption~\cite{max1995optical,levoy1988display,drebin1988volume};
cinematic rendering adds global illumination through path tracing~\cite{kroes2012exposure,comaniciu2016shaping,
dappa2016cinematic,eid2017cinematic,novak2018monte}. Neural and Gaussian representations have been optimised per
scan for visualisation~\cite{niedermayr2024application,kleinbeck2024multilayer,dyken2026veg,tang2025ivrgs,tang2026econgs,
weiss2022diffdvr,yao2025visnerf}, or combined with foveated path tracing~\cite{kleinbeck2026foveated}; a separate
line of work reconstructs CT from X-ray projections with
Gaussians~\cite{cai2024xgaussian,zha2024r2gaussian,gao2024ddgs}. We address feed-forward rendering of both CT and MR.

\noindent\textbf{TF editing and clipping.}
Designing the TF is the central interaction of volume viewing~\cite{kniss2002multidimensional,ljung2016state}.
Gaussians trained on images normally have one TF baked into their colours; VEG~\cite{dyken2026veg} and
FactorSplat~\cite{factorsplat} instead make colour and opacity a function of an editable TF on fixed geometry.
Clipping planes expose interior anatomy~\cite{weiskopf2003clipping}; ClipGS~\cite{li2025clipgs} learns clipping-aware
Gaussians, RaRa~\cite{li2025rara} approximates partially cut Gaussians, and XClipGS~\cite{xclipgs} cuts them exactly
in closed form.
Our \method reuses the TF model of FactorSplat and the clipping operator of XClipGS, and predicts the quantities they
need with a network instead of fitting them per scan.

\section{Method}
\label{sec:method}

\begin{figure*}[t]
\centering
\includegraphics[width=\textwidth]{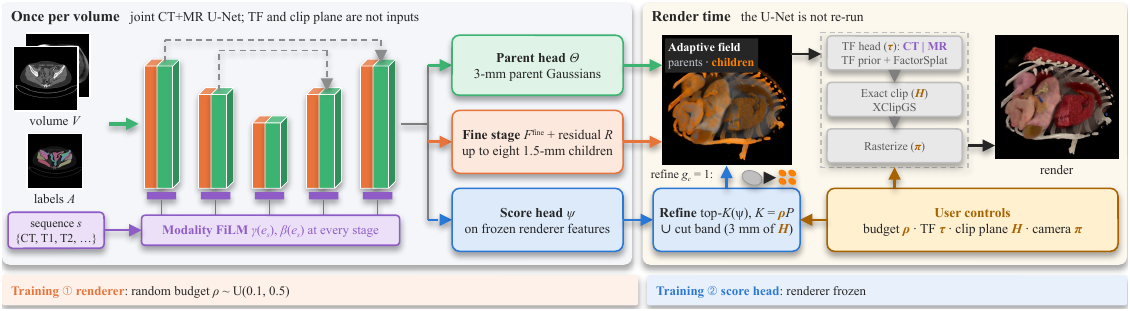}
\caption{\textbf{Overview of \method.} Once per volume, one U-Net for CT and MR predicts 3-mm parent Gaussians
(green), up to eight 1.5-mm children per parent (orange) and a score that ranks the parents (blue). At render time,
the top-$K$ parents under the budget $\rho$ and those next to the clipping plane $H$ are replaced by their children.
Bottom: training stages.}
\label{fig:pipeline}
\end{figure*}

\method works in two phases (\cref{fig:pipeline}). Once per volume, a network predicts Gaussians at two resolutions
and a score that tells where the finer ones are worth their cost. At render time, the budget and the clipping plane
decide which Gaussians are drawn, and a small network colours them for the current TF.
Nothing in the first phase depends on the TF or the clipping plane, so both can change at render time without a new
pass of the main network.

\subsection{Problem setup}
\label{sec:method:setup}
The input is a volume $V$, its anatomy label map $A$, and its acquisition type $s$, which is CT or the type of
the MR sequence (\eg T1 or T2).
The user controls three things: the camera $\pi$, the TF $\tau$, and an optional clipping half-space
$H=\{\mathbf{x}:\mathbf{n}^\top\mathbf{x}\le d\}$. A TF assigns a colour and an opacity to every intensity, with a
separate curve for each anatomy label.
We predict a set of Gaussians $\mathcal{G}$ whose rendering reproduces the image of a production path-traced volume
renderer for the chosen camera, TF and plane.
Two requirements guide the design. First, the geometry of $\mathcal{G}$ must not depend on $\tau$ or $H$, so that
editing the TF or moving the plane needs no new U-Net pass. Second, the number of rendered Gaussians must be
adjustable at render time, without retraining. We control it with a budget $\rho\in[0.1,0.5]$, the fraction of
coarse Gaussians that are refined (\cref{sec:method:children}).

\subsection{Network}
\label{sec:method:backbone}
We use the 3D U-Net~\cite{nnunet} of RenderFM~\cite{renderfm} and change its input and its output.
The input is the volume, resampled to 1.5-mm isotropic spacing, and a binary foreground mask ($A>0$). RenderFM also
gives the TF to the network; we do not, so one forward pass serves every TF.
The decoder outputs, for every cell $c$ of a 3-mm grid, the parameters $\Theta_c$ of one Gaussian and a 32-channel
feature vector $F_c$. We add one decoder stage that produces features $F^{\mathrm{fine}}$ on the 1.5-mm grid, which are used for the finer
Gaussians of \cref{sec:method:children}.

\noindent\textbf{Modality conditioning.}
CT and MR share the U-Net weights. To tell the network what it is looking at, we learn an embedding
$\mathbf{e}_s\in\mathbb{R}^{32}$ for each acquisition type. It scales and shifts the features $\mathbf{h}_\ell$ of
every encoder and decoder stage $\ell$ (FiLM~\cite{film}),
\begin{equation}
\mathbf{h}_\ell \leftarrow \mathbf{h}_\ell\odot\big(1+\boldsymbol{\gamma}_\ell(\mathbf{e}_s)\big) + \boldsymbol{\beta}_\ell(\mathbf{e}_s).
\end{equation}
The linear maps $\boldsymbol{\gamma}_\ell,\boldsymbol{\beta}_\ell$ are initialised to zero. The conditioning
therefore has no effect at first, and joint training starts exactly from a network pretrained on CT.

\noindent\textbf{Gaussian parameters.}
Each cell holds one directional Gaussian~\cite{gao2026direct}: a position at the cell centre, a scale, a rotation,
a colour (degree-1 spherical harmonics), an opacity, and a term that varies the opacity with the viewing direction.
As in RenderFM, colour and opacity start from the TF value at the cell's label and intensity, and the network
predicts a correction.
Some cells stay transparent under every TF. We skip a cell when no TF of the training set makes its label and
intensity visible. This test ignores the TF chosen at render time, so the Gaussians stay independent
of the TF.

\subsection{Coarse-to-fine refinement}
\label{sec:method:children}
A 3-mm Gaussian blurs structures that are thinner than its cell, such as a vessel wall or the boundary
between two organs. A 1.5-mm grid resolves them, but it has eight times as many cells, and most of them lie in uniform
tissue. We therefore keep the 3-mm grid and add finer Gaussians only where they improve the image.

We call the 3-mm Gaussians \emph{parents} and write $P$ for their number. The cell of parent $c$ contains eight
points $\mathbf{u}_{c,k}$ of the 1.5-mm grid, and each point can hold a finer Gaussian, a \emph{child} of $c$. The
visibility test above is applied to every child as well, so parent $c$ has $n_c\le 8$ children.
A child is predicted as its parent plus a correction,
\begin{equation}
\theta_{c,k} = \Theta_c + R\big(F^{\mathrm{fine}}(\mathbf{u}_{c,k}),\,F_c\big),
\end{equation}
where $R$ is a two-layer MLP that sees the fine features at the child's position and the features of the parent.
Its output layer is initialised to zero, so every child starts as a copy of its parent, moved to its own position
and halved in size. A child reads the label, intensity and TF value at its own position, so two siblings can show different tissues.

\noindent\textbf{Refinement replaces the parent.}
Children are not added on top of their parent; they replace it. Rendering both would count the same tissue twice,
and the children could only correct a blurred parent instead of representing the tissue themselves. Each parent has a switch $g_c\in\{0,1\}$. If
$g_c=1$, the parent is \emph{refined}: it is removed and its children are rendered instead,
\begin{equation}
\alpha_c \leftarrow (1-g_c)\,\alpha_c,\qquad \alpha_{c,k} \leftarrow g_c\,\alpha_{c,k},
\label{eq:compose}
\end{equation}
where $\alpha$ denotes opacity. Gaussians with zero opacity are dropped before rasterisation, so refining parent $c$
costs $n_c-1$ additional Gaussians. At $\rho{=}20\%$, the model renders about 2.5$\times$ as many Gaussians as with
parents only.

\noindent\textbf{Training with random budgets.}
We want one model that works at any budget and with any rule for choosing the refined parents. At every training
step we therefore draw a budget $\rho\sim\mathcal{U}(0.1,0.5)$ and refine $K=\mathrm{round}(\rho P)$ parents chosen
at random. The random order is fixed per scan, so the subsets of one scan at different budgets are nested.
The network thus learns to render any mixture of parents and children. The renderer is thus defined for any choice
of refined parents, which later allows us to estimate what refining one of them changes. Which parents \emph{should} be
refined is a separate question, answered in \cref{sec:method:score}.

\noindent\textbf{Cut band.}
A clipping plane exposes interior tissue that views from outside rarely show, and coarse Gaussians are most
noticeable on the cut face. Whenever a clipping plane is active, we therefore refine every parent whose centre lies
within 3\,mm of the plane, in addition to the $K$ parents of the budget. We call these parents the \emph{cut band}
and use it in training and at render time. The cut band adds Gaussians on clipped views, so we always report the
number of rendered Gaussians together with $\rho$.

\subsection{Learning where to refine}
\label{sec:method:score}
At render time we must decide which $K$ parents to refine. A random choice spends part of the budget where the
parents already suffice.
An obvious option is to learn the switches $g_c$ together with the renderer, but this lowered image quality in our
experiments (\cref{sec:exp:ablation}). We instead train the renderer first, with random switches as described above,
and learn afterwards where refinement helps.

\noindent\textbf{Estimating the benefit of refinement.}
With the renderer frozen, we ask for every parent how much the image error would drop if it were refined. Trying
the parents one by one is too expensive, so we estimate all of them at once with a gradient. We set every switch to
$g_c=0.5$, so that parents and children are both rendered at half opacity and the loss is differentiable with
respect to $g_c$. We then render a set $\mathcal{V}$ of training views and accumulate the negative gradient of the loss with
respect to every switch,
\begin{equation}
b_c = -\sum_{v\in\mathcal{V}} \frac{\partial \mathcal{L}_v}{\partial g_c}\Big|_{\mathbf{g}=0.5},
\qquad \mathcal{L}_v = 0.8\,\mathcal{L}_1 + 0.2\,(1-\mathrm{SSIM}).
\label{eq:benefit}
\end{equation}
To first order, $b_c$ is the decrease of the training loss $\mathcal{L}_v$ when parent $c$ is replaced by its
children.

\noindent\textbf{Score head.}
Computing $b_c$ needs reference images, which do not exist for a new volume. A small MLP $\psi$, the \emph{score
head}, therefore learns to predict $b_c$ from what is available at inference: the feature vector $F_c$ of the parent
and two statistics of its children, namely how much their intensities vary and whether they cross a label boundary.
These statistics indicate that the tissue inside the cell is not uniform, which is where a single Gaussian is too
coarse.
The scale of the benefit differs between scans, so the target is normalised per scan (\cref{sec:supp:impl}).
The score depends on neither the TF nor the view, so it is computed once per volume.
At render time we refine the $K=\mathrm{round}(\rho P)$ parents with the highest scores, together with the cut band
of the current clipping plane. A new budget only changes $K$, so it takes effect in the next frame without a U-Net pass.

\subsection{Appearance and clipping}
\label{sec:method:inherited}
The Gaussians described so far depend on neither the TF nor the clipping plane. Two components from prior work apply
both at render time.

\noindent\textbf{TF-dependent appearance.}
Colour and opacity under a TF $\tau$ follow FactorSplat~\cite{factorsplat}. Each Gaussian looks up $\tau$ at the
intensities it covers, and a small shared network, the \emph{TF head}, predicts from this lookup how its colour and
opacity respond. Gaussians that $\tau$ makes fully transparent
are removed. FactorSplat optimises the per-Gaussian
inputs of this head separately for each scene; here the U-Net predicts them, for parents and children alike. After a
TF edit, only the TF head is evaluated again.

The TF head is the only part of the model that is not shared between modalities: CT and MR each have their own copy,
while the U-Net and the score head are shared.
The MR copy is also given a simple physical rule as its starting point. Suppose a TF edit multiplies the TF opacity
of the tissue covered by a Gaussian by a factor $r$. If this tissue were a homogeneous medium whose density is
scaled by $r$, the opacity of the Gaussian would change from $o$ to
\begin{equation}
o_r = 1-(1-o)^{r}.
\end{equation}
The MR head starts from a blend of $o$ and $o_r$ and learns a correction (\cref{sec:supp:impl}).
The rule has no parameters and leaves images under the base TF unchanged. We use it only for MR; on CT it did not
improve validation results.

\noindent\textbf{Clipping.}
Clipping uses the closed-form operator of XClipGS~\cite{xclipgs}, which cuts every Gaussian exactly at the plane.
Without it, a plane could only keep or remove whole Gaussians, which leaves a ragged cut face.
Together with the cut band, the plane thus always meets fine Gaussians that are cut exactly.

\subsection{Training}
\label{sec:method:training}
Training has three steps. The first two train the renderer and the third trains the score head
(\cref{fig:pipeline}, bottom).
\emph{(1)~Parents.} The network is trained with parents only, first on CT and then on CT and MR together.
\emph{(2)~Children.} Children are switched on, and the model is trained on both modalities with random budgets and
the cut band. The two TF heads are also trained in this step. Since every child starts as a copy of its parent,
this step starts close to the result of step 1.
\emph{(3)~Score head.} The renderer is frozen and the score head is trained as described in \cref{sec:method:score}.
Steps 1 and 2 minimise $0.8\,\mathcal{L}_1+0.2\,(1-\mathrm{SSIM})$ between rendered and reference images at full
resolution. \Cref{sec:supp:impl} gives the schedule and the hyper-parameters.

\section{A CT+MR Rendering Benchmark}
\label{sec:benchmark}

\noindent\textbf{Scans.}
The CT part contains 1,008 scans from TotalSegmentator~\cite{totalsegmentator}: 895 for training, 46 for validation
and 67 for testing.
The MR part contains 958 scans (772 / 71 / 115) of the head, spine, knee, heart and abdomen. They come from
TotalSegmentator MRI~\cite{ds_totalsegmentator_mri}, IXI~\cite{ds_ixi}, AOMIC~\cite{ds_aomic},
SPIDER~\cite{ds_spider} and SKM-TEA~\cite{ds_skmtea}, with one scan per subject.
Where a source defines a test set, its scans stay in our test set.
For anatomy labels we use the reference labels of a source if it provides them and TotalSegmentator predictions
otherwise. \Cref{sec:supp:data} lists the splits per source.

\noindent\textbf{Reference images.}
The reference images are rendered by a production path tracer with denoising. The volumes are resampled to 0.75-mm
isotropic spacing and rendered at $1600^2$, the resolution used by RenderFM~\cite{renderfm}. The anatomy labels
are upsampled to the same spacing with smoothing, which avoids stair-step artefacts at label boundaries.

\noindent\textbf{Transfer functions and views.}
Every scan has a base TF, a set of clinical presets, and edits of these TFs, which change the opacity or colour of
all or selected structures. There are ten CT presets (\eg bone, lung, abdomen) and five reported MR presets
(anatomical, neuro cortex, target highlight, knee cartilage and abdominal organs).
A preset is generated from the intensity statistics of each scan and is used only if the structures it shows are
present.
Each view is rendered either without clipping or with an axis-aligned clipping plane.
A training scan has 128 views: half with the base TF and half with ten further TFs, a random selection of presets
and edits. Nearly half of the views are clipped. Validation and test scans use a fixed set of TFs and add views at unseen camera and plane positions.

\noindent\textbf{Reported families.}
We group the test conditions into four \emph{families}: the base TF without clipping, the base TF with clipping,
clinical presets, and presets with an additional edit (preset+edit).
They separate the three things a user changes: the view, the clipping plane and the TF.
Together they contain 6,269 CT and 9,400 MR test frames.
Training also uses TFs without a clinical counterpart, such as random edits. \Cref{sec:supp:data} specifies all TFs and
views.

\noindent\textbf{Metrics.}
We report PSNR, SSIM and LPIPS on full frames. We also report PSNR on cut faces only, \ie on the pixels where the
visible tissue lies on the clipping plane. Cut faces cover only a small part of a frame, so full-frame metrics
understate differences there.
Each metric is averaged over the frames of a patient and then over patients (for CT, which has no patient identifiers, over scans). Views with the base TF dominate the
frame count (91\% of MR frames), so we also report a family-balanced mean, the mean of the four family means.
When we compare two models, we compute the difference on the same frames and give a 95\% confidence interval from a
bootstrap over patients.

\section{Experiments}
\label{sec:exp}

\noindent\textbf{Baseline.}
We compare with RenderFM~\cite{renderfm}, the closest feed-forward method. Its published weights were trained with
a different reference renderer (21.15\,dB on our CT references), so we retrain RenderFM on our benchmark, starting
from the published weights and using as many updates as our CT pretraining. RenderFM is a CT method; for MR, we
fine-tune the retrained model on our MR training set. The baseline is thus one model per modality.
RenderFM has no clipping operator, so on clipped views we remove its Gaussians behind the plane.

\noindent\textbf{Variants of \method.}
To analyse our design we train two variants. The \emph{parents-only} model is trained in exactly the same way as \method but
without children. The \emph{specialists} are trained like \method but on CT only or on MR only; like the baseline,
they are one model per modality and serve as a reference.

Unless stated otherwise, \method uses $\rho{=}20\%$. Pairs of numbers are CT / MR, and brackets are 95\%
confidence intervals. \Cref{sec:supp:models} gives the training details of all models.
We first compare with RenderFM (\cref{sec:exp:main}). We then study refinement and the learned score
(\cref{sec:exp:ablation}), the effect of sharing one model between CT and MR (\cref{sec:exp:joint}), and the cost
at run time (\cref{sec:exp:cost}).

\subsection{Main comparison}
\label{sec:exp:main}
\begin{table*}[t]
\centering
\caption{\textbf{CT and MR results by family} (test: 67 CT, 115 MR patients; PSNR in dB, per-patient mean). All: all
frames; Fam.-bal.: mean of the four family means; \#G: Gaussians per view (millions); FPS: at $1600^2$
(\cref{tab:latency}). RenderFM: retrained, one model per modality. Parents only: \method without children (same \#G as
RenderFM). Bold: best single model. Grey: one model per modality.}
\label{tab:main}
\setlength{\tabcolsep}{3.5pt}
\footnotesize
\newcommand{\rf}[1]{\textcolor{gray}{#1}}
\begin{tabular}{@{}llcccccccccc@{}}
\toprule
 & & \multicolumn{3}{c}{All reported families} & \multicolumn{5}{c}{PSNR by family} & & \\
\cmidrule(lr){3-5}\cmidrule(lr){6-10}
Mod. & Method & PSNR$\uparrow$ & SSIM$\uparrow$ & LPIPS$\downarrow$ & \makecell{Base\\uncut} & \makecell{Base\\clipped}
 & \makecell{Clinical\\presets} & \makecell{Preset\\+ edit} & \makecell{Fam.-\\bal.} & \#G & FPS \\
\midrule
\multirow{4}{*}{CT} & RenderFM~\cite{renderfm} (retrained) & 26.87 & 0.936 & 0.085 & 26.91 & 27.25 & 25.72 & 26.32 & 26.55 & 0.35 & 99 \\
 & \method\ (parents only) & 29.39 & 0.948 & 0.068 & 29.96 & 29.69 & 27.32 & \textbf{28.98} & 28.99 & 0.35 & 225 \\
 & \textbf{\method} (joint) & \textbf{29.69} & \textbf{0.951} & \textbf{0.057} & \textbf{30.31} & \textbf{30.05} & \textbf{27.42} & 28.95 & \textbf{29.18} & 0.89 & 162 \\
 & \rf{\method\ (specialist)} & \rf{29.80} & \rf{0.952} & \rf{0.056} & \rf{30.41} & \rf{30.20} & \rf{27.53} & \rf{28.85} & \rf{29.25} & \rf{0.90} & \rf{163} \\
\midrule
\multirow{4}{*}{MR} & RenderFM~\cite{renderfm} (retrained) & 26.25 & 0.921 & 0.148 & 26.74 & 25.56 & 28.40 & 26.78 & 26.87 & 0.17 & 159 \\
 & \method\ (parents only) & 27.95 & 0.932 & 0.126 & 28.71 & 27.27 & 28.79 & 27.50 & 28.07 & 0.17 & 350 \\
 & \textbf{\method} (joint) & \textbf{28.63} & \textbf{0.939} & \textbf{0.107} & \textbf{29.23} & \textbf{28.11} & \textbf{29.31} & \textbf{28.13} & \textbf{28.70} & 0.42 & 229 \\
 & \rf{\method\ (specialist)} & \rf{28.93} & \rf{0.940} & \rf{0.107} & \rf{29.55} & \rf{28.38} & \rf{29.55} & \rf{28.44} & \rf{28.98} & \rf{0.41} & \rf{224} \\
\bottomrule
\end{tabular}
\end{table*}

\Cref{tab:main} shows the main results. Averaged over the reported families, \method reaches 29.69\,dB on CT and
28.63\,dB on MR, which is 2.82\,dB [+2.71, +2.91] and 2.38\,dB [+2.30, +2.45] above the retrained RenderFM.
SSIM and LPIPS improve as well. The improvement is largest on views with the base TF (3.4 / 2.5\,dB) and on
cut faces (2.1 / 2.6\,dB), and smallest on MR clinical presets (0.91\,dB [+0.78, +1.03]).
On presets with an additional edit, \method is 2.63 / 1.35\,dB above RenderFM. When the four
families are weighted equally, the gain is 2.63\,dB on CT and 1.83\,dB on MR.
RenderFM remains better on six of the 15 presets (glass, lung and skin on CT; abdominal organs, knee and neuro
cortex on MR).

\begin{figure*}[t]
\centering
\includegraphics[width=\textwidth]{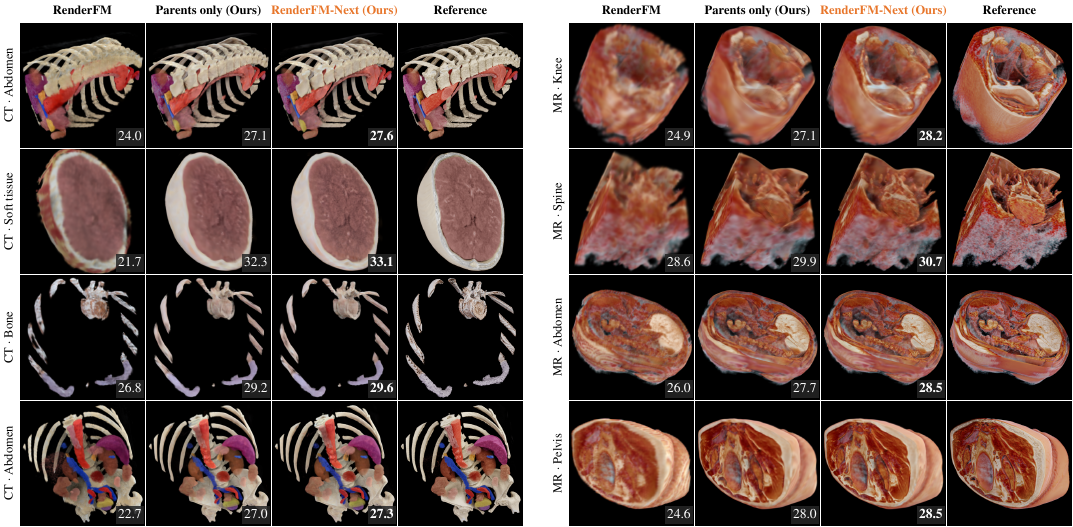}
\caption{\textbf{Qualitative comparison} on clipped clinical presets (validation; full-frame PSNR in dB, best in bold).
CT rows: abdomen, soft-tissue and bone presets; MR rows: anatomical preset on knee, spine, abdomen and pelvis scans.
Children
sharpen cut faces and thin structures that the parents-only model and RenderFM (retrained) blur.}
\label{fig:qualitative}
\end{figure*}

\Cref{fig:qualitative} shows examples on clipped clinical presets. RenderFM blurs the cut face and the boundaries
between organs. The parents-only model already renders them more faithfully, and the children sharpen cut faces and
thin structures further. \Cref{fig:supp:qual} shows typical frames.

Most of this improvement does not come from rendering more Gaussians. The parents-only model renders as many
Gaussians as RenderFM and is already better by 2.51\,dB [+2.42, +2.61] on CT and 1.69\,dB [+1.62, +1.76] on MR.
Refinement adds a further 0.30 / 0.69\,dB (\cref{sec:exp:ablation}).

\noindent\textbf{Compared with fitting each scan.}
A slower alternative to a feed-forward model is to fit Gaussians to each test scan with
FactorSplat~\cite{factorsplat}, which takes 4--23 minutes per scan. On six CT and six MR test patients, the fitted
models are 3.2\,dB worse than \method on CT and on par on MR (+0.09\,dB [$-$0.51, +0.66]). They are better on
unclipped views with the base TF, to which they were fitted (+0.7 / +2.3\,dB). They are worse on the clipped evaluation views ($-$1.8 / $-$1.0\,dB) and on CT presets outside their
training TFs ($-$3.8\,dB). A model trained on many patients is thus more robust on clipped views and on TFs outside
the fitting set, and it needs no fitting time.

\subsection{Effect of refinement}
\label{sec:exp:ablation}
\begin{table}[t]
\centering
\caption{\textbf{Effect of refinement.} PSNR differences (dB, 95\% CIs); \#G in millions.
\emph{Top}: relative to the parents-only model (test). \emph{Bottom}: learned selection at 20\% minus each
alternative (validation).}
\label{tab:children}
\setlength{\tabcolsep}{2.4pt}
\scriptsize
\begin{tabular}{@{}lcccc@{}}
\toprule
 & \multicolumn{2}{c}{CT} & \multicolumn{2}{c}{MR} \\
\cmidrule(lr){2-3}\cmidrule(l){4-5}
 & \#G & $\Delta$PSNR & \#G & $\Delta$PSNR \\
\midrule
\multicolumn{5}{@{}l}{\textit{Test, vs.\ the parents-only model (29.39 / 27.95\,dB)}} \\
Children switched off & 0.35 & $-$0.19 \pmci{$-$0.21}{$-$0.17} & 0.17 & $-$0.24 \pmci{$-$0.26}{$-$0.22} \\
\textbf{\method} & 0.89 & \textbf{+0.30} \pmci{+0.28}{+0.32} & 0.42 & \textbf{+0.69} \pmci{+0.66}{+0.71} \\
\midrule
\multicolumn{5}{@{}l}{\textit{Validation, learned 20\% minus alternative (\#G of the alternative)}} \\
Random selection & 0.82 & +0.18 \pmci{+0.16}{+0.21} & 0.49 & +0.18 \pmci{+0.16}{+0.21} \\
Label-boundary score & 0.74 & +0.09 \pmci{+0.07}{+0.12} & 0.48 & +0.07 \pmci{+0.06}{+0.09} \\
Gradient score & 0.88 & +0.09 \pmci{+0.05}{+0.12} & 0.49 & +0.14 \pmci{+0.12}{+0.17} \\
No cut band & 0.85 & +0.12 \pmci{+0.11}{+0.14} & 0.46 & +0.51 \pmci{+0.47}{+0.56} \\
All parents refined & 2.59 & +0.14 \pmci{+0.08}{+0.19} & 1.53 & +0.21 \pmci{+0.17}{+0.24} \\
\bottomrule
\end{tabular}
\end{table}

\begin{figure}[t]
\centering
\includegraphics[width=\linewidth]{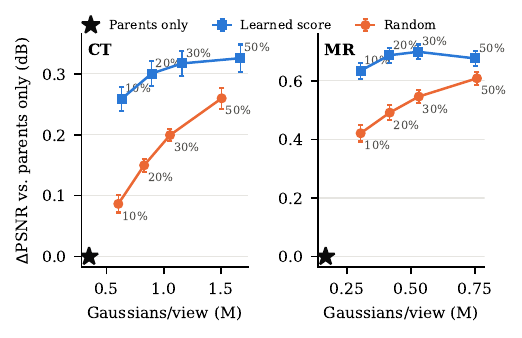}
\caption{\textbf{Quality vs.\ rendered Gaussians} (test). PSNR gain of \method over the parents-only model
(95\% CIs) with learned and with random selection at $\rho\in\{10,20,30,50\}\%$.}
\label{fig:quality_count}
\end{figure}

\noindent\textbf{Does refinement help?}
We compare \method with the parents-only model, which differs only in that it never uses children
(\cref{tab:children}, top). Refinement improves PSNR by 0.30\,dB on CT and 0.69\,dB on MR.
The gain is largest where coarse Gaussians are most visible: 0.37 / 0.84\,dB on clipped views and
0.51 / 1.17\,dB on cut faces. It is smaller on clinical presets (0.10 / 0.52\,dB), and on CT preset+edit
refinement brings no gain ($-$0.03\,dB [$-$0.08, +0.01]).
When the children of \method are switched off at render time, it falls 0.19 / 0.24\,dB below the parents-only
model. The gain therefore comes from the children and not from better parents.

\noindent\textbf{Does the learned score help?}
\Cref{fig:quality_count} compares selection by the learned score with random selection on the test set.
The learned score is better at every budget on both modalities. At $\rho{=}20\%$ it gains 0.30 / 0.69\,dB over
the parents-only model, against 0.15 / 0.49\,dB for random selection with about as many Gaussians.
Refining a learned 10\% of the parents matches refining a random 50\%. On CT the difference is $-$0.00\,dB
[$-$0.02, +0.02] with 2.4$\times$ fewer rendered Gaussians (0.63 vs.\ 1.51\,M), and on MR it is +0.03\,dB
[+0.01, +0.04] (0.30 vs.\ 0.76\,M).
A small budget placed well is thus worth as much as a large one placed at random, which keeps refinement
affordable at render time.
The learned curve flattens early: from 20\% to 50\%, PSNR changes by only +0.03 / $-$0.01\,dB while the number of
Gaussians nearly doubles. The default $\rho{=}20\%$ was chosen on validation data; \cref{sec:supp:results}
lists the values behind the figure and the frame time at every budget.

On validation, we also compare the learned score with two hand-designed scores, one based on label boundaries and
texture and one on image-space gradients as in AbsGS~\cite{absgs} (\cref{tab:children}, bottom). The learned score
is better than both. It is also better than refining every parent, which renders three times as many Gaussians and
lies outside the budgets seen in training.

\noindent\textbf{Cut band.}
Without the cut band, validation PSNR drops by 0.12 / 0.51\,dB overall and by 0.26 / 0.94\,dB on clipped views.
A wider band of 4.5\,mm adds only 0.02 / 0.04\,dB, so we keep 3\,mm, the spacing of the parents.

\noindent\textbf{Why not learn the selection with the renderer?}
In a CT-only pilot we learned the switches $g_c$ jointly with the renderer through the rendering loss. The
resulting model was 0.44\,dB worse than a parents-only model, whereas training with random budgets and refining a
random 30\% of the parents was 0.15\,dB better (\cref{sec:supp:results}). We therefore learn the score after the renderer.

\subsection{One model for CT and MR}
\label{sec:exp:joint}
The grey rows of \cref{tab:main} show the specialists. They use the same architecture and training recipe as the
joint model, with the same amount of training per modality, and start from the same network pretrained on CT.
Like RenderFM, they are one model per modality, and they are 2.93 / 2.68\,dB above it.
Using one model for both modalities costs 0.11\,dB [0.08, 0.14] on CT and 0.29\,dB [0.28, 0.31] on MR and halves
the total number of parameters.
On MR the cost is similar in every family (0.24--0.32\,dB). On CT it is small, and on CT preset+edit the joint
model is slightly better (+0.10\,dB [+0.02, +0.19]).
In return for this cost, a single set of weights serves both modalities.

\begin{table}[t]
\centering
\caption{\textbf{Results per data source} (test; PSNR in dB, all reported families; $n$: patients). Top row: CT;
other rows: MR. Grey: one model per modality.}
\label{tab:mrsources}
\setlength{\tabcolsep}{2.0pt}
\footnotesize
\newcommand{\rf}[1]{\textcolor{gray}{#1}}
\begin{tabular}{@{}llccccc@{}}
\toprule
Source & Region & $n$ & RenderFM & \makecell{Parents\\only} & \textbf{\method} & \rf{Spec.} \\
\midrule
TotalSeg.\ CT & various & 67 & 26.87 & 29.39 & \textbf{29.69} & \rf{29.80} \\
\midrule
TotalSeg.\ MRI & various & 23 & 25.60 & 27.18 & \textbf{27.78} & \rf{28.02} \\
IXI & head & 32 & 26.72 & 28.60 & \textbf{29.35} & \rf{29.68} \\
AOMIC & head & 19 & 27.80 & 29.56 & \textbf{30.39} & \rf{30.74} \\
SPIDER & spine & 5 & 25.83 & 27.28 & \textbf{27.92} & \rf{28.20} \\
SKM-TEA & knee & 36 & 25.49 & 27.09 & \textbf{27.72} & \rf{27.97} \\
\bottomrule
\end{tabular}
\end{table}

All CT scans come from one source, whereas the MR test set combines five sources that differ in body region,
scanner and resolution.
\Cref{tab:mrsources} shows that the MR results do not depend on any one of them. In every MR source, \method is
2.1--2.6\,dB above RenderFM, 0.6--0.8\,dB above the parents-only model, and 0.24--0.35\,dB below the MR
specialist. It is also above RenderFM on clinical presets and on preset+edit in every source (\cref{tab:supp:mrsources}).

\subsection{Cost}
\label{sec:exp:cost}
An interactive viewer has three kinds of cost: preparing a new volume, reacting to a TF edit, and rendering a
view. \Cref{tab:latency} compares \method with RenderFM on all three. All times are measured on one
A100 GPU at $1600^2$.

\noindent\textbf{Rendering and editing.}
With parents only, \method is cheaper than RenderFM in every row. After a TF edit, RenderFM has to run its U-Net
again, while \method only evaluates the TF head: the edit takes 21 / 15\,ms with parents only and 51 / 35\,ms at
$\rho{=}20\%$, against 71 / 34\,ms for RenderFM.
At $\rho{=}20\%$, \method renders about 2.5$\times$ more Gaussians (\cref{tab:main}). It still renders more than
160 views per second (6.2 / 4.4\,ms per view), but refinement has a cost. Preparing a new volume takes
1.73 / 1.06\,s, about twice as long as with parents only and still less than RenderFM (1.99 / 1.56\,s). Peak
memory more than doubles, from 3.0 / 1.5\,GB to 7.7 / 4.9\,GB. Preparation time and memory do not depend on the
budget, but rendering time does: between $\rho{=}10\%$ and $30\%$, a view takes 5.3 to 7.0\,ms on CT
(3.7 to 4.9\,ms on MR), and the gain over the parents-only model goes from 0.26 to 0.32\,dB (0.63 to 0.70\,dB).
Moving a clipping plane takes 17 / 12\,ms, because only the cut band changes.

\noindent\textbf{Stability under camera motion.}
Interactive viewing also requires that the image does not flicker when the camera moves.
On camera orbits of seven test patients we measure how closely the change between consecutive frames
follows that of the reference (\cref{sec:supp:results}).
\method reaches 33.1 / 32.8\,dB and RenderFM 31.7 / 31.8\,dB.

\section{Future Work}
\label{sec:future}
The Gaussians of \method store the appearance under the fixed lighting of the reference renderer. A natural next
step is relighting: an appearance that responds to changes of lighting and to tissue removed by clipping.

\section{Conclusion}
\label{sec:conclusion}
\method predicts Gaussians for CT and MR volumes at two resolutions and lets the user decide at render time how
much of the volume is rendered at the finer one. Training the renderer with random refinement budgets, and learning
afterwards where refinement pays off, gives a model that improves over the same model without refinement on both
modalities. It improves over a retrained RenderFM by 2.8\,dB on CT and 2.4\,dB on MR, and it renders a $1600^2$ view in 4--6\,ms. The learned score makes refinement efficient: refining
a learned 10\% of the coarse Gaussians matches refining a random 50\% with 2.4$\times$ fewer Gaussians. The budget can be changed at render time without retraining or a new network pass. TF editing and exact clipping are supported through components reused from prior work.
The accompanying benchmark covers clinical presets, TF edits and clipping for both modalities and reports
differences with confidence intervals over patients.

\section*{LLM Usage Statement}
We used generative AI models (Anthropic Claude and OpenAI Codex) throughout this work: to help write and debug
code, to run and analyse experiments, and to draft and edit the text, tables and figures of this paper and its
appendix. The authors designed the method and the experiments, checked all results and all text, and take full
responsibility for the content.
{
    \small
    \bibliographystyle{ieeenat_fullname}
    \bibliography{main}
}

\maketitlesupplementary
\setcounter{table}{0}\setcounter{figure}{0}
\renewcommand{\theHtable}{supp.\arabic{table}}\renewcommand{\theHfigure}{supp.\arabic{figure}}
\appendix

\noindent This supplement gives the details that the main paper refers to.
\Cref{sec:supp:data} describes the benchmark: sources and splits, reference volumes, TFs and views.
\Cref{sec:supp:impl} completes the description of the model, its training schedule and the rendering implementation.
\Cref{sec:supp:models} states how the compared models were trained.
\Cref{sec:supp:results} contains additional results: per-family, per-preset and per-source tables, the budget sweep,
checks of the learned score, the pilot on learning the switches with the renderer, the per-scan fit, temporal
stability, typical frames and a description of the video.

\section{Benchmark details}
\label{sec:supp:data}

\begin{table}[t]
\centering
\caption{\textbf{Benchmark sources and splits} (scans). CT excludes 84 thick-slice scans; MR excludes one degenerate
scan.}
\label{tab:supp:sources}
\setlength{\tabcolsep}{3pt}
\footnotesize
\begin{tabular}{@{}llrrr@{}}
\toprule
Source & Anatomy / sequence & Train & Val. & Test \\
\midrule
CT: TotalSegmentator~\cite{totalsegmentator} & whole body & 895 & 46 & 67 \\
\midrule
MR: TotalSegmentator MRI~\cite{ds_totalsegmentator_mri} & body, mixed & 187 & 17 & 23 \\
MR: IXI~\cite{ds_ixi} & head, T1/T2 & 278 & 25 & 32 \\
MR: AOMIC~\cite{ds_aomic} & head, T1 & 166 & 15 & 19 \\
MR: SPIDER~\cite{ds_spider} & lumbar spine, T2 & 32 & 4 & 5 \\
MR: SKM-TEA~\cite{ds_skmtea} & knee, qDESS & 109 & 10 & 36 \\
\bottomrule
\end{tabular}
\end{table}

\noindent\textbf{Sources and splits.}
\Cref{tab:supp:sources} lists the scans per source and split. We exclude 84 thick-slice CT scans and one degenerate
MR scan. Where a source defines a test set (TotalSegmentator MRI and SKM-TEA), we keep it. The remaining validation
and test scans are stratified by source and, where available, by anatomy or site.
Every MR subject contributes one scan: the identifiers of IXI, AOMIC and SPIDER are per subject, and for SKM-TEA we
use one time point per patient. A search of the MR volumes found no duplicate or near-duplicate pair.

\noindent\textbf{Statistics.}
\Cref{tab:supp:data} summarizes the scans per source. The sources differ by more than an order of magnitude in two
respects: the size of the working grid (median 7.8\,M voxels for SKM-TEA knees vs.\ 106.2\,M for CT) and the number
of anatomy labels present (median 3 for SPIDER vs.\ 74 for CT). Both determine the number of parents and children.

\begin{table*}[t]
\centering
\caption{\textbf{Dataset statistics} per source (all splits; median [IQR] over scans). Spacing: smallest / largest native
voxel spacing; FOV: longest extent; grid: working grid (superscript: scans at 1.0 instead of 0.75\,mm); labels: anatomy
labels present; parents: 3-mm grid; \#G: rendered per view at $\rho{=}20\%$ (test). $^\ast$CT release resampled to 1.5\,mm.}
\label{tab:supp:data}
\setlength{\tabcolsep}{2.2pt}
\scriptsize
\begin{tabular}{@{}lrcccccccc@{}}
\toprule
Source & Scans & \makecell{In-plane\\spacing (mm)} & \makecell{Slice\\spacing (mm)} & FOV (mm) & \makecell{Grid\\(M voxels)} & Labels
 & \makecell{Parents\\(M)} & \makecell{Cand.\\children (M)} & \makecell{\#G, test\\(M)} \\
\midrule
CT: TotalSegmentator & 1008 & 1.5$^\ast$ & 1.5$^\ast$ & 405 [332, 572] & 106.2 [53.7, 192.2] & 74 [52, 100] & 0.43 [0.22, 0.62] & 3.17 [1.64, 4.58] & 0.92 [0.54, 1.16] \\
\midrule
MR: TotalSeg.\ MRI & 227 & 1.09 [0.73, 1.25] & 2.50 [1.50, 3.00] & 380 [256, 400] & 49.3 [22.4, 66.9]$^{9}$ & 25 [6, 29] & 0.39 [0.15, 0.64] & 3.08 [1.16, 5.07] & 0.95 [0.34, 1.38] \\
MR: IXI & 335 & 0.94 [0.94, 0.94] & 1.20 [1.20, 1.20] & 240 [240, 240] & 24.5 [24.5, 24.7] & 6 [6, 6] & 0.15 [0.14, 0.17] & 1.21 [1.10, 1.31] & 0.39 [0.33, 0.42] \\
MR: AOMIC & 200 & 1.00 [1.00, 1.00] & 1.00 [1.00, 1.00] & 256 [256, 256] & 24.8 [24.8, 24.8] & 6 [5, 6] & 0.12 [0.11, 0.13] & 0.95 [0.90, 1.04] & 0.30 [0.28, 0.33] \\
MR: SPIDER & 41 & 0.47 [0.47, 0.47] & 0.90 [0.90, 0.90] & 300 [300, 300] & 18.4 [18.4, 18.7] & 3 [3, 3] & 0.13 [0.12, 0.14] & 0.99 [0.91, 1.12] & 0.30 [0.29, 0.31] \\
MR: SKM-TEA & 155 & 0.31 [0.31, 0.31] & 0.80 [0.80, 0.80] & 160 [160, 160] & 7.8 [7.8, 7.8] & 9 [8, 9] & 0.06 [0.05, 0.06] & 0.43 [0.40, 0.48] & 0.14 [0.13, 0.15] \\
\bottomrule
\end{tabular}
\end{table*}

\noindent\textbf{Labels.}
CT uses the 117 TotalSegmentator classes plus a body label. MR uses the 50 TotalSegmentator MRI classes plus 13
derived or dataset-specific labels (body, spinal canal, ventricles, knee bones, cartilage and menisci).
Where a source has no reference MR segmentation, we predict the labels with TotalSegmentator
MRI~\cite{ds_totalsegmentator_mri}. On MR, the body label marks unlabelled tissue inside the body outline.

\noindent\textbf{Reference volumes.}
Volumes are resampled to 0.75-mm isotropic spacing. Nine TotalSegmentator MRI scans, in which an axis would
otherwise exceed 610 voxels, are resampled to 1.0\,mm instead.
Labels are upsampled one at a time: each label is smoothed with a Gaussian of $\sigma{=}1$ fine voxel, and every
voxel takes the label with the largest value.
This avoids the stair-step artefacts that nearest-neighbour 1.5-mm labels produce at $1600^2$. The network sees these labels subsampled to its 1.5-mm grid.

\noindent\textbf{Transfer functions.}
\emph{Presets.} The ten clinical CT presets are soft tissue, abdomen, lung, bone, CTA, cardiac, kidney, glass, skin
and target. The five clinical MR presets are anatomical, neuro cortex, target highlight, MSK knee (the knee-cartilage
preset of the main paper) and abdominal organs. Presets are generated from the
intensity statistics of each scan (Hounsfield units for CT, per-label quantiles for MR). A preset is used for a
scan only if its key structures are present, \eg a target label for the target preset or a contrast phase for CTA.

\emph{Edits.} The edit families are global random, translucency, multi-region, focus, isolate and hide.

\emph{Training TFs.} Each training scan has its base TF and ten further TFs, drawn with a seed that is fixed per
scan: four eligible presets, two presets with an edit, and edits of the base TF for the rest. A training TF must
show at least 30\% of the labelled anatomy, and presets that show more are drawn more often.

\emph{Evaluation TFs.} The evaluation set is fixed: the base TF, the eligible presets and the preset+edit
compositions. A preset+edit composition applies one edit from inside the training edit range to a preset. On CT, these are the
bone preset with a +100 shift of the opacity level and the glass preset with a change of hue and opacity. On MR, it
is the anatomical preset with a +300 shift (0--4095 axis).

\noindent\textbf{Views and clipping.}
CT and MR training scans use the same design: 64 views with the base TF (28 uncut, 36 clipped) and 64 views with
the other TFs (4 uncut per TF and 24 clipped). Cameras use a 60\textdegree{} perspective, and the uncut directions
cover the full sphere. Clipping planes are axis-aligned and cut at 35, 45, 55 or 65\% of the volume extent.
Validation and test scans add novel views with the base TF, 4 uncut and 6 clipped at new plane positions of
40--60\%, and one uncut and one clipped view per eligible preset and preset+edit composition (one clipped CT
test view of the lung preset is missing).

\begin{table}[t]
\centering
\caption{\textbf{Evaluation frames}. Top: patients / frames per reported family (validation has no MR
preset+edit frames). Bottom: patients eligible for each reported preset (val. / test).}
\label{tab:supp:frames}
\setlength{\tabcolsep}{2pt}
\scriptsize
\begin{tabular}{@{}llccccc@{}}
\toprule
Mod. & Split & Base uncut & Base clipped & Clinical presets & Preset + edit & All \\
\midrule
CT & Val. & 46 / 1{,}472 & 46 / 1{,}932 & 46 / 724 & 46 / 184 & 46 / 4{,}312 \\
 & Test & 67 / 2{,}144 & 67 / 2{,}814 & 67 / 1{,}043 & 67 / 268 & 67 / 6{,}269 \\
\midrule
MR & Val. & 71 / 2{,}272 & 71 / 2{,}982 & 71 / 394 & -- & 71 / 5{,}648 \\
 & Test & 115 / 3{,}680 & 115 / 4{,}830 & 115 / 660 & 115 / 230 & 115 / 9{,}400 \\
\bottomrule
\end{tabular}\\[4pt]
\begin{tabular}{@{}lrr@{\hspace{10pt}}lrr@{}}
\toprule
CT preset & Val. & Test & MR preset & Val. & Test \\
\midrule
abdomen & 35 & 50 & abdominal organs & 10 & 13 \\
bone & 46 & 67 & anatomical & 71 & 115 \\
cardiac & 35 & 43 & MSK knee & 10 & 36 \\
CTA & 14 & 24 & neuro cortex & 42 & 55 \\
glass & 46 & 67 & target highlight & 64 & 111 \\
kidney & 12 & 27 &  & &  \\
lung & 36 & 43 &  & &  \\
skin & 46 & 67 &  & &  \\
soft tissue & 46 & 67 &  & &  \\
target & 46 & 67 &  & &  \\
\bottomrule
\end{tabular}
\end{table}

\noindent\textbf{Evaluation frames and metrics.}
\Cref{tab:supp:frames} lists the frames behind every reported number. There are 6,269 CT and 9,400 MR test frames
and 4,312 / 5,648 validation frames; the MR preset+edit composition has no validation frames. The mean of a preset
is taken over the patients for which it is eligible (\eg 13 of 115 MR test patients for abdominal organs).
Besides full-frame PSNR, SSIM and LPIPS, the tables below report two local metrics. Edge PSNR is computed within
4\,px of strong gradients of the reference image. Cut-face PSNR is computed on TF-visible tissue whose first
visible surface is the clipping plane, eroded by 2\,px.

\section{Method details}
\label{sec:supp:impl}

Below, a \emph{candidate} is a parent with at least one child.

\begin{table}[t]
\centering
\caption{\textbf{Training stages} of the final model. The first two rows form step~1 of the main paper (parents)
and the last row step~2 (children). Views: rendered views per update.}
\label{tab:supp:train}
\setlength{\tabcolsep}{5pt}
\footnotesize
\begin{tabular}{@{}lrlrr@{}}
\toprule
Stage & Steps & LR & Views & GPU-h \\
\midrule
CT pretraining & 32k & $10^{-4}$ & 32 & 47 \\
Joint CT+MR & 12k\,+\,3k & $5\!\times\!10^{-5}$ & 16, 8 & 17 \\
Children & 8.4k & $5\!\times\!10^{-5}$ & 20 & 21 \\
\bottomrule
\end{tabular}
\end{table}
\begin{table*}[t]
\centering
\caption{\textbf{Per-family metrics} for the models of \mainref{tab:main} (test; patient means over the reported
families). Edge: PSNR within 4\,px of strong reference gradients. Cut: cut-face PSNR over all clipped reported views.
Grey: one model per modality.}
\label{tab:supp:families}
\setlength{\tabcolsep}{2.6pt}
\footnotesize
\begin{tabular}{@{}llccccccccccccc@{}}
\toprule
 & & \multicolumn{3}{c}{Base uncut} & \multicolumn{3}{c}{Base clipped} & \multicolumn{3}{c}{Clinical presets} & \multicolumn{3}{c}{Preset + edit} & \\
\cmidrule(lr){3-5}\cmidrule(lr){6-8}\cmidrule(lr){9-11}\cmidrule(lr){12-14}
Mod. & Method & SSIM & LPIPS & Edge & SSIM & LPIPS & Edge & SSIM & LPIPS & Edge & SSIM & LPIPS & Edge & Cut \\
\midrule
CT & RenderFM & 0.933 & 0.088 & 17.81 & 0.939 & 0.081 & 17.51 & 0.933 & 0.090 & 15.86 & 0.935 & 0.085 & 16.84 & 19.36 \\
 & \textbf{\method\ (Ours)} & 0.951 & 0.057 & 20.88 & 0.955 & 0.050 & 20.24 & 0.943 & 0.074 & 17.50 & 0.951 & 0.073 & 18.89 & 21.46 \\
 & \textcolor{gray}{Specialist (Ours)} & \textcolor{gray}{0.951} & \textcolor{gray}{0.056} & \textcolor{gray}{20.96} & \textcolor{gray}{0.955} & \textcolor{gray}{0.049} & \textcolor{gray}{20.35} & \textcolor{gray}{0.943} & \textcolor{gray}{0.074} & \textcolor{gray}{17.54} & \textcolor{gray}{0.952} & \textcolor{gray}{0.072} & \textcolor{gray}{18.89} & \textcolor{gray}{21.54} \\
\midrule
MR & RenderFM & 0.929 & 0.150 & 17.14 & 0.911 & 0.151 & 17.07 & 0.947 & 0.106 & 17.13 & 0.931 & 0.142 & 17.46 & 18.64 \\
 & \textbf{\method\ (Ours)} & 0.943 & 0.126 & 19.22 & 0.933 & 0.095 & 19.47 & 0.955 & 0.085 & 17.76 & 0.944 & 0.110 & 19.03 & 21.24 \\
 & \textcolor{gray}{Specialist (Ours)} & \textcolor{gray}{0.944} & \textcolor{gray}{0.126} & \textcolor{gray}{19.52} & \textcolor{gray}{0.934} & \textcolor{gray}{0.094} & \textcolor{gray}{19.73} & \textcolor{gray}{0.955} & \textcolor{gray}{0.084} & \textcolor{gray}{17.94} & \textcolor{gray}{0.945} & \textcolor{gray}{0.110} & \textcolor{gray}{19.22} & \textcolor{gray}{21.42} \\
\bottomrule
\end{tabular}
\end{table*}

\noindent\textbf{Network and Gaussians.}
The U-Net decoder stops at stride 2. It outputs a 29-channel parameter field and a 32-channel feature map on the
3-mm grid. One added stage, a transposed convolution followed by a convolution block with the full-resolution skip,
yields the 1.5-mm features. The directional Gaussian~\cite{gao2026direct} stores a mean view direction and a
$3\times3$ Cholesky precision factor, which together modulate the opacity with the viewing direction. Children use
half the base scale of their parent and a softer opacity prior (prior exponent $\times0.5$).

\noindent\textbf{TF heads.}
The TF head follows FactorSplat~\cite{factorsplat}. Its inputs are a 64-bin lookup of the TF over the intensities
in the footprint of each Gaussian, rank-8 per-Gaussian factors and a 16-channel anatomy code; the last two are
predicted by the U-Net. An exact visibility test removes the Gaussians that the TF makes invisible.
The head is duplicated at the start of the children stage. From then on, the CT copy is trained on CT scans and the
MR copy on MR scans. The MR copy starts halfway between $o$ and $o_r$ in logit space ($o_r$ as defined in the main
paper) and adds its learned opacity response at half scale.
The ratio $r$ is computed per Gaussian from the same lookup. Let $\bar a_\tau$ be the mean TF opacity over the voxels
in the footprint of the Gaussian, with hidden labels counted as zero. Then $r=\bar a_\tau/\bar a_{\tau_0}$, where
$\tau_0$ is the base TF; we cap $r$ at 20. A footprint that mixes several tissues thus receives one average ratio, and
the learned response corrects for the difference. Gaussians that are almost transparent under the base TF
($\bar a_{\tau_0}<10^{-3}$) have no ratio; they keep $o$ and rely on the learned response alone.

\noindent\textbf{Random budgets.}
In the children stage, the candidates of a scan are put in a random order that is seeded by the scan identifier and
kept fixed. The refined subset is the first $K=\mathrm{round}(\rho P)$ candidates of this order, so the subsets of
one scan at different budgets are nested. When fewer than $K$ candidates exist, all of them are refined.

\noindent\textbf{Score head.}
The benefit $b_c$ is heavy-tailed and its scale differs between scans, so the head regresses a squashed per-scan
z-score,
$t_c=\mathrm{sigmoid}\big(z_{\text{scan}}(\ln(1+10\,[b_c]_+/q_{90}))\big)$, where $q_{90}$ is the scan's 90th percentile
of $|b_c|$. The benefit is measured at $g_c=0.5$, because at $g_c=0$ the children are removed and receive no
gradient. The head is a LayerNorm--Linear(34$\to$64)--SiLU--Linear(64$\to$1) MLP. Its inputs are the detached
parent feature, the log standard deviation of the children's intensities and a label-boundary flag. It is trained
with an L1 loss between $\mathrm{sigmoid}(\psi_c)$ and $t_c$ for 4k steps with AdamW (learning rate $10^{-3}$,
cosine decay).
The labels come from 120 CT and 120 MR training scans with 16 views each; 216 are used for training and 24 held out.

\noindent\textbf{Hand-designed scores.}
The two hand-designed scores of \mainref{tab:children} rank the same candidates as the learned score and are used
in the same way: the top 20\% are refined, together with the cut band. The label-boundary score is
$100\,\beta_c+\ln(1+\sigma_c)$. Here $\beta_c\in\{0,1\}$ marks a parent with a child whose label differs from that
of the parent or of a neighbouring voxel, and $\sigma_c$ is the standard deviation of the intensities of its
children. Parents on a label boundary thus come first, ordered by texture. The gradient score follows the
densification criterion of AbsGS~\cite{absgs}. We render eight evaluation views of the scan with parents only and
sum, for every parent, the magnitude of the gradient of the loss with respect to its position, converted to pixels
(multiplied by depth over focal length). This score needs the reference images of the scan, which a new volume
does not have, so it is a reference point and not a rule that could be deployed.

\noindent\textbf{Training schedule.}
\Cref{tab:supp:train} lists the stages: CT pretraining, joint CT+MR training, in which MR scans are sampled more
often in the last 3k steps, and children training with random budgets. The score head is trained afterwards with the
renderer frozen.
The three renderer stages use AdamW (weight decay $10^{-5}$), linear warm-up followed by polynomial decay (power 0.9), mixed
precision, and 4 views of one scan per GPU and step. The loss is
$0.8\,\mathcal{L}_1+0.2\,(1-\mathrm{SSIM})$ at $1600^2$.

\noindent\textbf{Rendering implementation.}
Rendering has three kinds of cost (\mainref{tab:latency}): preparing a new volume, updating after a TF edit, and
rendering a view, possibly after the clipping plane has moved.

\emph{Preparation.} The geometry of a volume is computed once and stored as a compact plan: the cropped volume
and labels, the 3-mm parent grid, and the candidates with their eligible children. When the plan is loaded, the
network inputs and the positions and opacity priors of the Gaussians are derived on the GPU. All candidate
children are prepared, whatever the budget. Preparation at $\rho{=}20\%$ takes 1.7 / 1.1\,s on CT / MR, against
0.8 / 0.5\,s with parents only; the difference comes mainly from preparing the children.

\emph{TF edit.} A TF edit re-evaluates only the TF head, which takes 51 / 35\,ms at $\rho{=}20\%$
(\mainref{tab:latency}). At the first TF, only the rendered Gaussians are decoded:
the parents that are not replaced and the children of the top-$K$ parents. Cut-band children are decoded when a
clipping plane first needs them.

\emph{Cached TF.} The output of the TF head for a TF does not depend on the view, apart from a final
view-dependent opacity factor. It is therefore computed once and cached, and further views with the same TF
apply only this factor and rasterize. Switching back to a cached TF does not run the TF head: it adds 0.25\,ms
before rasterization on CT and MR (0.22\,ms with parents only), so the next frame is rendered at the usual frame
time.

\emph{Clip-plane move.} Moving the plane recomputes the cut band and decodes the children that are newly needed.
The visibility and the TF appearance of every decoded Gaussian are cached per TF, so the TF head runs only on the
newly decoded Gaussians. At $\rho{=}20\%$, a move takes 17 / 12\,ms (median, CT / MR).

\begin{figure*}[t]
\centering
\includegraphics[width=\textwidth]{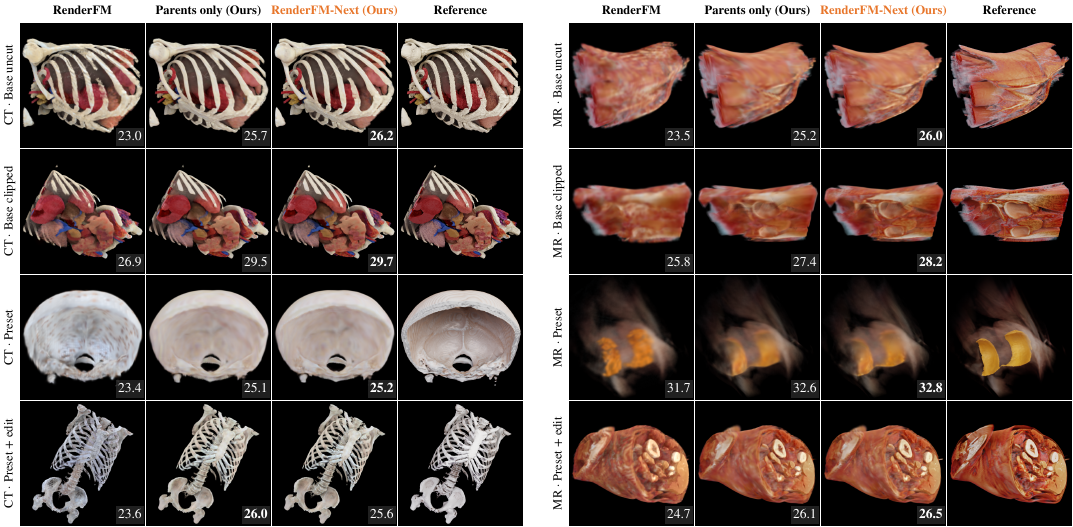}
\caption{\textbf{Representative views}, one per reported family (validation; full-frame PSNR in dB, best in bold).
Each example is the frame whose PSNR gain over retrained RenderFM is the median of its family.}
\label{fig:supp:qual}
\end{figure*}

\section{Compared models}
\label{sec:supp:models}

\noindent\textbf{Differences from RenderFM.}
\mainref{tab:latency} lists what \method changes with respect to RenderFM. With parents only, \method renders as
many Gaussians as RenderFM, scores higher (\mainref{tab:main}) and is cheaper in every row. Children add quality at the cost
reported there.

\noindent\textbf{RenderFM, retrained.}
We retrain RenderFM on CT for 32k updates from its published weights, as many as in our CT pretraining. For MR,
we fine-tune the CT-retrained weights on the MR training set. The fine-tuning uses the same number of rendered MR
training views as our MR-only model (168k), the same MR preset weighting in the matching stages, and the per-scan
MR intensity normalisation of \method. TF edits use the TF input channels and the opacity prior of the original
work. RenderFM has no exact clipping operator, so on clipped views we remove the Gaussians whose mean lies
behind the plane.

\noindent\textbf{RenderFM, published weights.}
Without retraining, the published weights score 21.15\,dB on our CT test references. They were trained with the
reference renderer of the original work, whose references differ from ours in TF palette, label smoothing and
lighting. Our implementation reproduces the original inference code on our references (within 0.3\,dB per view).
All quality comparisons in the paper use the retrained model.

\noindent\textbf{Parents-only model.}
The parents-only model starts from the same initialisation as the children stage and is trained with the same
number of steps, views per update, sampling, TF heads and checkpoint rule. It never uses children.

\noindent\textbf{Specialists.}
A specialist is trained like the joint model, with the same architecture, children, score head, default budget and
cut band, but on one modality only. Each specialist receives as much training on its modality as the joint model
gives to that modality, counted in rendered views, in the parents stage and in the children stage. It has one TF
head (the MR specialist with the opacity rule $o_r$) and its own score head. The joint model and both specialists
start from the same CT-pretrained network.

\section{Additional results}
\label{sec:supp:results}

\noindent\textbf{Per family and per preset.}
\Cref{tab:supp:families} gives SSIM, LPIPS, edge PSNR and cut-face PSNR per family, and \cref{tab:supp:presets}
PSNR per clinical preset, for the models of \mainref{tab:main}. \method is better than retrained RenderFM in every
family and metric. Per preset, RenderFM is ahead on glass, lung and skin (CT) and on abdominal organs, MSK knee
and neuro cortex (MR).

\begin{table}[t]
\centering
\caption{\textbf{Per-preset PSNR} (dB, test, patient means; $n$: patients with that preset) for the models of
\mainref{tab:main}. Bold: best single model. Grey: one model per modality.}
\label{tab:supp:presets}
\setlength{\tabcolsep}{1.8pt}
\footnotesize
\begin{tabular}{@{}llrccc@{}}
\toprule
Mod. & Preset & $n$ & RenderFM & \makecell{\textbf{\method}\\\textbf{(Ours)}} & \textcolor{gray}{\makecell{Specialist\\(Ours)}} \\
\midrule
CT & abdomen & 50 & 26.40 & \textbf{29.37} & \textcolor{gray}{29.19} \\
CT & bone & 67 & 25.30 & \textbf{27.17} & \textcolor{gray}{26.69} \\
CT & cardiac & 43 & 33.48 & \textbf{33.74} & \textcolor{gray}{33.75} \\
CT & CTA & 24 & 31.82 & \textbf{35.86} & \textcolor{gray}{35.90} \\
CT & glass & 67 & \textbf{29.39} & 24.75 & \textcolor{gray}{24.79} \\
CT & kidney & 27 & 25.56 & \textbf{30.47} & \textcolor{gray}{30.57} \\
CT & lung & 43 & \textbf{33.20} & 32.21 & \textcolor{gray}{32.56} \\
CT & skin & 67 & \textbf{22.79} & 21.18 & \textcolor{gray}{21.30} \\
CT & soft tissue & 67 & 22.04 & \textbf{29.52} & \textcolor{gray}{29.60} \\
CT & target & 67 & 18.27 & \textbf{23.32} & \textcolor{gray}{24.28} \\
\midrule
MR & abd. organs & 13 & \textbf{28.04} & 26.88 & \textcolor{gray}{26.97} \\
MR & anatomical & 115 & 26.25 & \textbf{28.82} & \textcolor{gray}{29.13} \\
MR & MSK knee & 36 & \textbf{28.29} & 26.84 & \textcolor{gray}{27.19} \\
MR & neuro cortex & 55 & \textbf{30.10} & 29.44 & \textcolor{gray}{29.64} \\
MR & target highlight & 111 & 30.08 & \textbf{30.98} & \textcolor{gray}{31.15} \\
\bottomrule
\end{tabular}
\end{table}

\noindent\textbf{Per data source.}
\Cref{tab:supp:mrsources} splits the MR results of \mainref{tab:mrsources} by family.
The order on All PSNR is the same in every source (specialist, joint model, parents-only model, RenderFM), and the
joint model trails the specialist by 0.24--0.35\,dB. Retrained RenderFM is below the joint model on clinical
presets and preset+edit in every source. CT comes from one source and is not split further.

\begin{table}[t]
\centering
\caption{\textbf{MR test PSNR per source and family} (dB, patient means); All as in \mainref{tab:mrsources}.
$n$: test patients; every MR test patient has preset and preset+edit frames. Grey: one model per modality.}
\label{tab:supp:mrsources}
\setlength{\tabcolsep}{2.2pt}
\scriptsize
\newcommand{\rf}[1]{\textcolor{gray}{#1}}
\begin{tabular}{@{}llccccc@{}}
\toprule
Source & Method & All & \makecell{Base\\uncut} & \makecell{Base\\clipped} & \makecell{Clin.\\presets} & \makecell{Preset\\+ edit} \\
\midrule
\multirow{4}{*}{\makecell[l]{TotalSeg.\ MRI\\($n{=}23$)}} & RenderFM & 25.60 & 25.41 & 25.48 & 27.43 & 26.35 \\
 & Parents only (Ours) & 27.18 & 27.26 & 27.04 & 27.77 & 27.26 \\
 & \textbf{\method\ (Ours)} & 27.78 & 27.72 & 27.75 & 28.30 & 27.82 \\
 & \rf{Specialist (Ours)} & \rf{28.02} & \rf{28.03} & \rf{27.95} & \rf{28.55} & \rf{27.99} \\
\midrule
\multirow{4}{*}{\makecell[l]{IXI\\($n{=}32$)}} & RenderFM & 26.72 & 27.82 & 25.60 & 28.76 & 26.69 \\
 & Parents only (Ours) & 28.60 & 30.05 & 27.47 & 29.18 & 27.48 \\
 & \textbf{\method\ (Ours)} & 29.35 & 30.42 & 28.53 & 29.74 & 28.15 \\
 & \rf{Specialist (Ours)} & \rf{29.68} & \rf{30.74} & \rf{28.89} & \rf{29.95} & \rf{28.56} \\
\midrule
\multirow{4}{*}{\makecell[l]{AOMIC\\($n{=}19$)}} & RenderFM & 27.80 & 28.40 & 27.02 & 29.77 & 28.56 \\
 & Parents only (Ours) & 29.56 & 30.42 & 28.73 & 30.81 & 29.27 \\
 & \textbf{\method\ (Ours)} & 30.39 & 31.17 & 29.69 & 31.31 & 29.92 \\
 & \rf{Specialist (Ours)} & \rf{30.74} & \rf{31.50} & \rf{30.08} & \rf{31.50} & \rf{30.25} \\
\midrule
\multirow{4}{*}{\makecell[l]{SPIDER\\($n{=}5$)}} & RenderFM & 25.83 & 25.77 & 25.66 & 27.62 & 26.72 \\
 & Parents only (Ours) & 27.28 & 27.35 & 27.10 & 28.75 & 27.06 \\
 & \textbf{\method\ (Ours)} & 27.92 & 27.89 & 27.82 & 29.43 & 27.67 \\
 & \rf{Specialist (Ours)} & \rf{28.20} & \rf{28.20} & \rf{28.08} & \rf{29.65} & \rf{27.59} \\
\midrule
\multirow{4}{*}{\makecell[l]{SKM-TEA\\($n{=}36$)}} & RenderFM & 25.49 & 25.90 & 24.78 & 28.09 & 26.19 \\
 & Parents only (Ours) & 27.09 & 27.73 & 26.49 & 28.03 & 26.79 \\
 & \textbf{\method\ (Ours)} & 27.72 & 28.31 & 27.17 & 28.49 & 27.43 \\
 & \rf{Specialist (Ours)} & \rf{27.97} & \rf{28.64} & \rf{27.36} & \rf{28.79} & \rf{27.79} \\
\bottomrule
\end{tabular}
\end{table}

\noindent\textbf{Budget sweep.}
\Cref{tab:supp:budget} lists the values behind \mainref{fig:quality_count} (confidence intervals are shown there) and adds the frame times. Learned selection beats random
selection at every budget. On CT the learned gain grows from +0.26\,dB at 10\% to +0.33\,dB at 50\%; on MR it is
flat (+0.63 to +0.70\,dB). From 10\% to 30\% the frame time grows from 5.3 to 7.0\,ms on CT and from 3.7 to
4.9\,ms on MR; 50\% costs 8.7 / 6.0\,ms.
The default budget was selected on validation by a rule fixed beforehand: per modality, the smallest budget whose
validation PSNR is within 0.03\,dB of the best budget. It selected $\rho{=}20\%$ for CT and MR.

\begin{table}[t]
\centering
\caption{\textbf{Budget sweep} (test): the values of \mainref{fig:quality_count}. $\Delta$: PSNR gain (dB) over
the parents-only model; \#G: rendered Gaussians (M); ms: frame time with learned selection. Parents-only model
(CT / MR): 29.39\,dB, 0.35\,M, 4.4\,ms / 27.95\,dB, 0.17\,M, 2.9\,ms.}
\label{tab:supp:budget}
\setlength{\tabcolsep}{3.2pt}
\footnotesize
\begin{tabular}{@{}lcccccccccc@{}}
\toprule
 & \multicolumn{5}{c}{CT} & \multicolumn{5}{c}{MR} \\
\cmidrule(lr){2-6}\cmidrule(l){7-11}
 & \multicolumn{2}{c}{Learned} & \multicolumn{2}{c}{Random} & & \multicolumn{2}{c}{Learned} & \multicolumn{2}{c}{Random} & \\
\cmidrule(lr){2-3}\cmidrule(lr){4-5}\cmidrule(lr){7-8}\cmidrule(lr){9-10}
$\rho$ & $\Delta$ & \#G & $\Delta$ & \#G & ms & $\Delta$ & \#G & $\Delta$ & \#G & ms \\
\midrule
10\% & +0.26 & 0.63 & +0.09 & 0.60 & 5.3 & +0.63 & 0.30 & +0.42 & 0.30 & 3.7 \\
20\% & +0.30 & 0.89 & +0.15 & 0.83 & 6.2 & +0.69 & 0.42 & +0.49 & 0.42 & 4.4 \\
30\% & +0.32 & 1.16 & +0.20 & 1.06 & 7.0 & +0.70 & 0.53 & +0.55 & 0.53 & 4.9 \\
50\% & +0.33 & 1.67 & +0.26 & 1.51 & 8.7 & +0.68 & 0.75 & +0.61 & 0.76 & 6.0 \\
\bottomrule
\end{tabular}
\end{table}

\noindent\textbf{Selection at matched Gaussians.}
At the same budget, the selection rules of \mainref{tab:children} do not render exactly the same number of
Gaussians on CT (0.74 to 0.88\,M per view). We therefore repeat the comparison on validation with random selection
and the two hand-designed scores, where each rule refines as many parents as it needs to render as many Gaussians
as the learned score at $\rho{=}20\%$ (within 1\%). The learned score remains better overall: by 0.17 / 0.19\,dB
over random selection, by 0.06 / 0.08\,dB over the label-boundary score and by 0.09 / 0.14\,dB over the gradient
score (all intervals exclude zero).

\noindent\textbf{Estimated and measured benefit.}
We check the benefit estimate $b_c$ of \mainref{eq:benefit} against a direct measurement. On 24 validation scans
per modality, we take 240 parents per scan, spread evenly over the range of the learned score. For each of them we
render eight views twice, with this parent unrefined and refined and all other parents as at render time
($\rho{=}20\%$ and the cut band). The measured benefit $D_c$ is the loss of the first rendering minus that of the
second, and $b_c$ is computed on the same views. The estimate follows the measurement: the rank correlation
between $b_c$ and $D_c$ is 0.66 [0.63, 0.69] on CT and 0.61 [0.56, 0.65] on MR, and both have the same sign for
77\% / 75\% of the parents (means over scans, bootstrap over scans).

\noindent\textbf{Learning the switches with the renderer (CT pilot).}
Before building the joint model, we compared two ways of deciding which parents are refined. Both were 4k-step
continuations of the CT-pretrained model, and each was compared with a parents-only continuation trained
identically (0.35\,M Gaussians). Training with random budgets and
refining a random 30\% gains 0.15\,dB [+0.14, +0.15] at 1.06\,M Gaussians. Learning the switches through the
rendering loss with a 30\% budget loses 0.44\,dB [$-$0.48, $-$0.40] at 1.11\,M. We therefore did not use learned switches in the joint model.

\noindent\textbf{Per-scan fit.}
We fit FactorSplat to each of 6 CT and 6 MR test patients, with our renderer and losses, on the 128 training
views of the scan (base TF and 10 training TFs). We then score the fit, \method and retrained RenderFM on the same
evaluation frames (\cref{tab:supp:perscan}).
On CT the fit is 3.21\,dB below \method [$-3.48$, $-2.89$] and worse on all six patients. On MR the two are on
par ($+0.09$\,dB [$-0.51$, $+0.66$]).
The fit is better where it was fitted: on uncut views with the base TF (+0.7 / +2.3\,dB), and also on MR
preset+edit (+1.9\,dB). It is worse on the clipped evaluation views ($-$1.8 / $-$1.0\,dB) and on CT clinical
presets outside its training TFs ($-$3.8\,dB; on MR $-$0.9\,dB, not significant).
Fitting takes 18 / 7\,min per patient (CT / MR, mean). The fit renders 1.14 / 0.47\,M Gaussians per view, against
0.85 / 0.41\,M for \method.
Presets and edits are 85\% of these frames, so the All column differs from \mainref{tab:main}.

\begin{table}[t]
\centering
\caption{\textbf{Per-scan fit vs.\ feed-forward} (6 CT / 6 MR test patients, identical frames; PSNR in dB).
Cut: cut-face PSNR; Fit: per-scan optimisation per patient (minutes, A100).}
\label{tab:supp:perscan}
\setlength{\tabcolsep}{1.8pt}
\scriptsize
\begin{tabular}{@{}llccccccc@{}}
\toprule
 & Method & All & Uncut & Clipped & Presets & P+edit & Cut & Fit \\
\midrule
\multirow{3}{*}{CT} & Per-scan fit & 25.34 & \textbf{30.70} & 27.68 & 24.08 & 21.80 & 17.02 & 18\,m \\
 & RenderFM & 26.13 & 26.37 & 26.50 & 26.04 & 25.76 & 17.98 & -- \\
 & \textbf{\method\ (Ours)} & \textbf{28.55} & 29.99 & \textbf{29.50} & \textbf{27.91} & \textbf{28.34} & \textbf{20.22} & -- \\
\midrule
\multirow{3}{*}{MR} & Per-scan fit & \textbf{29.11} & \textbf{31.45} & 27.29 & 29.02 & \textbf{30.56} & 19.63 & 7\,m \\
 & RenderFM & 27.15 & 26.68 & 25.79 & 28.95 & 27.23 & 19.33 & -- \\
 & \textbf{\method\ (Ours)} & 29.02 & 29.14 & \textbf{28.31} & \textbf{29.95} & 28.61 & \textbf{21.51} & -- \\
\bottomrule
\end{tabular}
\end{table}

\noindent\textbf{Temporal stability.}
We render 48-frame camera orbits (0.75$^\circ$ steps) of 7 test patients with base, preset and clipped views, and
compare the frame-to-frame changes with those of the converged reference (tPSNR). \method reaches 33.1 / 32.8\,dB
(CT / MR) and RenderFM 31.7 / 31.8\,dB.

\noindent\textbf{Representative views.}
\Cref{fig:supp:qual} shows typical frames, selected by a fixed rule. They are taken from a pool of validation scans: all 46 on CT and 31 on MR, with ten base-TF views and the reported
preset frames of each scan. The reported MR preset+edit composition has no validation frames, so for this figure
we rendered it for the 31 MR scans. For each modality and family we rank the frames of the pool by their PSNR gain
over retrained RenderFM and show the median frame. On base TFs \method is better than RenderFM in almost every
frame of the pool (458 of 460 on CT, 309 of 310 on MR), and children add detail on cut faces. On presets and edits the typical gain
is smaller, in line with \mainref{tab:main}.

\noindent\textbf{Video.}
The video on the project page (70\,s) shows one whole-body CT scan and one whole-body MR scan of the test set side by
side, rendered by \method at the default budget ($\rho{=}20\%$ and the cut band). One prediction per scan serves
all frames, and neither scan is fitted. After a short overview of the pipeline, the video has three parts: a
camera orbit with the base TF; a sequence of TF switches (CT: bone, glass, CTA, kidney, target and abdomen; MR:
abdominal organs, target highlight and four edits: a translucent body and isolated organs, vessels or bone), in which only the
TF head is evaluated and the U-Net is not run again; and a coronal and a sagittal clipping plane that move through
the volume. The frame rates shown in the video are the measured values of \mainref{tab:latency}.

\end{document}